# Information fusion in the immune system

Jamie Twycross *, Uwe Aickelin

School of Computer Science, University of Nottingham, Nottingham NG8 1BB, UK

a b s t r a c t

Biologically-inspired methods such as evolutionary algorithms and neural networks are proving useful in the field of information fusion. Artificial immune systems (AISs) are a biologically-inspired approach which take inspiration from the biological immune system. Interestingly, recent research has shown how AISs which use multi-level information sources as input data can be used to build effective algo-rithms for realtime computer intrusion detection. This research is based on biological information fusion mechanisms used by the human immune system and as such might be of interest to the information fusion community. The aim of this paper is to present a summary of some of the biological information fusion mechanisms seen in the human immune system, and of how these mechanisms have been implemented as AISs.

## 1. Introduction

There is an increasing interest within the field of multi-sensor data fusion in biologically-inspired methods such as evolutionary algorithms [27] and neural networks [16]. The field of artificial immune systems (AISs) is an emerging biologically-inspired method which builds systems based on algorithms inspired by the biological immune system. AIS research has provided a number of general purpose techniques and algorithms which have successfully been applied to a range of optimisation, classification and data mining problems. As with evolutionary algorithms and neural networks, AISs could also provide useful solutions to optimisation and classification problems in multi-sensor data fusion.

More interestingly though perhaps, recent research in AISs [14,15,35,36] shows the importance of multi-level information in the construction of AISs. New models for AISs are emerging that are inspired by research in immunology into the role of the innate immune system in overall immune system dynamics. These AISs, which incorporate mechanisms inspired by both the innate and adaptive immune systems, are called second generation AISs. They stand in contrast to first generation AISs, which are inspired by adaptive immune system mechanisms only. One of the conse- quences of incorporating innate and adaptive mechanisms, as well as one of the defining characteristics of second generation AISs, is the need for a multi-level problem representation, and a multi-le- vel interaction of the components of the AIS with the problem [36].

As systems that integrate multi-level information sources, second generation AISs share much in common with multi-sensor data fusion systems. In this sense, researchers within the fields of AISs and multi-sensor data fusion have the potential to benefit from each other's findings. This paper focusses on the integration of multi-level information in AISs. The first section gives a brief introduction to AISs, and is followed by a short overview of biological mechanisms of information fusion seen in the human immune system. An implementational framework which allows AISs to be built which model these mechanisms is then summarised. An algorithm inspired by biological information fusion seen in the immune system is then presented, along with results from a number of experiments. This paper concludes with a discussion of the role of multi-level information sources in AISs.

## 2. Artificial immune systems

The field of artificial immune systems began in the early 1990s with a number of independent groups conducting research which used the biological immune system as inspiration for solutions to problems in other domains. There are several general reviews of AIS research [1,5,17], and a number of books including [6,8,32] covering the field. Large bibliographies have been collated by Das- gupta and Azeem [7] (over 600 journal and conference papers) and an annual international conference, ICARIS [30], has been held since 2002.

AISs can be broadly divided into two categories based on the mechanisms they implement: network-based models and population-based models [8], although this distinction is blurred as many hybrid models also exist. The first of these categories refers to systems that are largely based on idiotypic networks. Idiotypic networks are networks which model interactions between antibodies and antibodies as well as between antibodies and antigens. Population-based models use negative or clonal selection as the method of generating and maintaining a population of detectors. Generally, population-based models begin with the pseudo-random generation of a population of detectors. Negative selection refers to the removal of detectors which match instances in a training set. Clonal selection refers to the expansion and refinement of detectors which match instances.

* Corresponding author.
E-mail address: jpt@cs.nott.ac.uk (J. Twycross).

AISs have been built for a wide range of application domains including document classification, clustering, optimisation, fraud detection, and network- and host-based intrusion detection. On benchmark datasets, AISs have been shown to offer comparable and in some cases better performance compared to existing statis- tical and machine learning techniques. In particular, AISs may offer advantages over traditional approaches in problem domains such as dynamic clustering and classification, where data are continu- ously gathered and incorporated into existing clusters or classes, which themselves change over time [33]. AISs may also offer advantages over traditional algorithms in the classification of large static datasets. Many standard classification techniques are not amenable to parallelisation, whereas distributed immune-based classification algorithms have been developed [37], allowing large amounts of data to be efficiently processed in parallel.

In [17], Hart and Timmis assess the current state of AIS research. They point to the solid base of research which now exists using AISs to model the biological immune system, solve artificial or benchmark problems, and produce solutions to real-world applica- tions. At the same time they highlight the somewhat scattergun approach which has been taken in the field to date, with naive met- aphors often applied to problems that other approaches have al- ready tackled with some success. However, less research exists that addresses what the necessary components and organisation of AISs might be from a more general systems perspective. The ap- proach to date has generally been one of applying novel algorithms to existing problems.

We believe that the current state of AIS is understandable when one considers the biological basis on which much of it has been based: the mechanisms of the adaptive immune system. The focus of AIS research on the adaptive immune system has been in some ways similar to Artificial Intelligence's early concentration on the human mind and symbolic information processing. Only more re- cently has the scope of AI been widened by the acknowledgement of intelligence in the wider sense of adaptive behaviour of organ- isms other than humans.

Second generation AISs represent a new approach in AIS. They show that considering the biological immune system as composed of interacting innate and adaptive subsystems can be a profitable model of reality for AISs. A number of general design principles, de- tailed in [34], can be applied to build second generation AISs. Such AISs employ multi-level information sources as input data for pop- ulations of artificial cells or agents, which process and integrate this information. Such AISs can be used as recognition, control and monitoring systems [14,15,35,36].

Of particular interest to the information fusion community could be the mechanisms that are employed by the biological im- mune system to combine information from a variety of different sources. Essentially, the innate and adaptive immune systems sense different aspects of the state of an organism, and interact to combine this information to provide a robust and accurate mon- itoring and control system. Our research has developed a number of biologically-inspired algorithms, one of which is described in this paper, which combine information from a number of different sources.

The advantage of our second generation AIS algorithms is that they are able to correlate data from multiple noisy sensors, even in the presence of unknown time delays. For example, in both bio- logical systems and computer systems there is often a time delay between an event (such as infection by a biological or computer virus, respectively) and the consequences of this event (malfunctioning of the biological or computer system). A priori we do not know how long this delay may be. If we knew this we could prob- ably use an existing static machine learning algorithm. However, because such information is unavailable, we need a new type of algorithm.

Furthermore, noise is an inherent factor in both biological sen- sors and the sensors employed in our algorithms. If a single infor- mation source could be used to deduce accurate predictions, then there would be little point in using other information sources. However, no single indicator of system state has been found to ex- ist in many biological and artificial systems. Instead, the data from each information source provides a noisy and partial picture of the overall state of the system, and information from a number of dif- ferent information sources needs to be combined to determine an accurate picture of the system. Our approach offers a way of com- bining information from a large number of noisy and inaccurate sensors. As sensing is population-based and distributed across a large number of simple sensors, our approach should also be ro- bust against damage to individual sensors.

## 3. The human immune system

Biological systems have provided the inspiration for a number of novel biologically-inspired computational approaches such as genetic algorithms and neural networks. The first step in building effective biologically-inspired systems is an understanding of the biological system from which inspiration is drawn. This under- standing of how biology solves the problems which nature poses can then be mapped to artificial systems. In this section current understanding of the human immune system, which forms the ba- sis and justification for the algorithms described in the remainder of this paper, is briefly reviewed.

The human body is an amazingly complex organism which can be viewed at a number of levels. Cells are the basic structural and functional units of biological organisms. All together, humans have around $10^{14}$ cells. Cells are able to interact with their environment and communicate and coordinate their behaviour with other cells by synthesising and responding to a range of molecules. Molecules in the immediate environment of a cell are sensed by receptor proteins which are bound to the outer surface of the cell. These receptors can be thought of as locks, which are activated when a specific molecule, called the ligand (or key), binds to the receptor. Activation of the receptor initiates changes in the metabolism and function of the cell.

As well as receiving signals via receptors, cells also synthesise molecules that are ligands for receptors on other cells. These sig- nalling molecules can either be membrane-bound, in which case direct (cognate) contact between cells is necessary for receptor activation, or they can be released into the environment of the cell. Secreted molecules which mediate and regulate cell behaviour are called cytokines, and molecules which stimulate cell movement are called chemokines. Cells within the body aggregate to form tissue, such as muscle or connective tissue. Tissues themselves combine to form organs, such as the heart, brain, or thymus. Groups of these organs work together tightly to form systems, such as the cardio- vascular system or immune system [2,25].

Structurally, the immune system is a collection of cells, mole- cules, tissue, organs and circulatory systems [20]. Immune system cells are produced and mature in specialised areas of the body called primary lymphoid organs such as the thymus or bone mar- row. They are transported via the cardiovascular and lymphatic cir- culatory systems to peripheral tissues or specialised secondary lymphoid organs such as the lymph nodes or spleen. The body it- self exists in a world which is full of microorganisms. Many of these microorganisms find the body a rich resource of energy and material, and, if left unchecked, would consume so much of these resources and cause such damage to the body that its destruction and death would occur. Damage to the body is called pathology, and the damaging agent, such as a bacteria or virus, a pathogen. Functionally, the human immune system is able to locate and remove many of these pathogens from the body and maintain the body in a healthy state for many years.

This view of the immune system we have just described, one of a multi-level dynamic system of cells, molecules, tissue, organs and circulatory systems, is important for AISs. It provides the basis for a representation of second generation AISs as systems of autonomous agents which exist within a distributed and compartmentalised environment. These agents interact with each other and their environment through models of receptors, ligands and intercellular signalling. This mechanism of interaction is key to the dynamics of the biological system and relies on multi-level sources of information.

### 3.1. Innate and adaptive immunity

The immune system is often divided into two distinct yet interrelated subsystems: the innate immune system and adaptive immune system. The innate immune system is characterised as having three roles: host defence in the early stages of infection through non-specific recognition of a pathogen, induction of the adaptive immune response, and determination of the type of adap- tive response [18]. The main characteristics of adaptive immunity are specific recognition of pathogen leading to the generation of pathogen-specific long-term memory [20].

Differences between the innate and adaptive immune systems can be seen on a number of levels and are summarised in Table 1. The adaptive immune system is organised around two classes of cells: T cells and B cells, while the classes of cells of the innate immune system are much more numerous, including natural killer (NK) cells, dendritic cells (DCs), and macrophages. Cells within these classes are further subdivided into different types, such as naive or helper T cell, or immature, semimature or mature DC.

The receptors of innate system cells are entirely germline-encoded, in other words their structure is determined by the genome of the cell and has a fixed, genetically-determined specificity. Adaptive immune system cells possess somatically generated variable-region receptors such as T and B cell receptors with varying specificities, created by a complex process of gene segment rearrangement within the cell. On a population level, this leads to a non-clonal distribution of receptors on innate immune system cells, meaning that all cells of the same type have receptors with identical specificities. Receptors on adaptive immune system cells, however, are distributed clonally in that there are subpopulations of a specific cell type (clones) which all possess receptors with identical specificities, but that generally, cells of the same type have receptors with different specificities [18–20,28,29].

**Table 1**
Differences between innate and adaptive immunity.

| Property | Innate immune system | Adaptive immune system |
|---|---|---|
| Cells | DC, NK, macrophage | T cell, B cell |
| Receptors | Germline-encoded | Encoded in gene segments |
| | No somatic rearrangement | Somatic rearrangement necessary |
| | Non-clonal distribution | Clonal distribution |
| Recognition | Conserved molecular patterns | Details of molecular structure |
| | Selected over evolutionary time | Selected over lifetime of individual |
| Response | Cytokines, chemokines | Clonal expansion, cytokines |
| Action time | Immediate effector activation | Delayed effector activation |
| Evolution | Vertebrates and invertebrates | Only vertebrates |

The variable-region receptors of the adaptive immune system respond to features of pathogen structure, with B cell receptors di- rectly recognising peptide sequences on pathogens, such as com- ponents of bacterial cell membranes, and T cell receptors recognising peptide sequences. These receptors are selected for over the lifetime of the organism by processes such as clonal expansion, deletion or anergy and are under adaptive not evolu- tionary pressure. The immune system utilises adaptation of vari- able-region receptors to keep pace with evolutionary more rapid pathogens. This involves processes of cell selection such as clonal expansion, deletion and anergy, which take several days [19,29].

Conversely, innate immune system receptors recognise a genetically-determined set of ligands under evolutionary pressure. One key group of innate receptors is the pattern recognition receptor superfamily which recognises evolutionary-conserved pathogen-associate molecular patterns. Pattern recognition receptors do not recognise a specific feature of a specific pathogen as variable-region receptors do, but instead recognise common features or products of an entire class of pathogens. Thus, innate immune system receptors are termed non-specific, while adaptive immune system receptors are termed specific. The Toll-like receptor (TLR) family of pattern recognition receptors is the best characterised, and most mammals have around 10 to 15 different TLRs. For example, TLR4 is activated by lipopolysaccharide (LPS), a major component in the cell membrane of all gram-negative bacteria, TLR5 is activated by flagellin, a protein that forms the flagellum used by many classes of bacteria for locomotion, and TLR9 by unmethylated DNA found in DNA viruses [31].

Dendritic cells (DCs) of the innate immune system lie at the heart of the generation of peripheral tolerance. Tolerance is the ability of the immune system to react in a non-biodestructive manner to stimuli and has long been associated with adaptive immunity. Tolerance is usually discussed in terms of apoptosis or anergy of self-reactive T and B cells, and was initially proposed to occur centrally in a relatively short perinatal period, as epitomised in the clonal selection theory of Burnet [4]. While recent research shows the continuing importance of central tolerance mechanism [11,22], it is now accepted that peripheral tolerance mechanisms which operate to censor cells throughout the lifetime of the host are of equal importance.

### 3.2. Discussion

As outlined in the previous section, the biological immune system can be seen to carry out information fusion in a particular manner. The cells of the immune system, through their different repertoires of receptors, sense different levels of information relating to the state of the tissues of the body. Variable-region receptors on adaptive immune system cells sense structural features, i.e. the protein composition of cells in the tissue. Variable-region receptors can be produced to recognise any possible protein sequence, and the particular set of proteins sequences that the immune system can recognise at any one time is determined by on-line learning mechanisms over the lifetime of the individual. On-line learning mechanisms are still an under-researched area in multi-sensor data fusion [38], and the biological immune system could provide an important source of inspiration for the development of such mechanisms.

In contrast, the germline-encoded receptors of innate immune system cells respond to behavioural as well as structural features of the cells which make up and inhabit the tissues of the body. By behavioural features of cells we mean what the cell is doing
i.e. what proteins it is producing, as opposed to the proteins which the cell is composed of. Having access to behavioural features provides immune system cells with a different level of information concerning the tissues of the body. Also, as detailed above, certain innate immune system receptors respond to protein sequences that are common to entire classes of pathogens. Thus, adaptive and innate immune system cells together can be seen to sense information at several different levels: the proteins that structure individual cells; the proteins that structure classes of cells; and the proteins that are produced by cell.

What is more, the biological immune system combines this multi-level information in a decentralised and distributed manner, fusing the information from individual immune system cells. There is no centralised controller in the biological immune system. Decentralised and distributed data fusion has a number of advantages over centralised fusion: robustness, scalability, survivability and modularity [10,24]. The biological immune system could provide a rich source of inspiration in the development of decentralised and distributed information fusion systems. Multi-agent information fusion systems such as [13,26] are currently an active area of research. As well as the different levels of information sensed by the immune system, its overall organisation and the mechanisms of control and communication that exist could be used to develop more sophisticated multi-agent information fusion systems.

## 4. System overview

The aim of this section is to summarise the implementation of libtissue, a prototype software system for building second generation AISs and applying them to real-world problems. The libtissue software allows researchers to implement AISs as multi-agent systems and analyse the behaviour of these systems when they are applied to real-world problems. In particular, libtissue is designed to allow researchers to implement second generation AISs.

The libtissue system has a client/server architecture as pictured in Fig. 1. An AIS is implemented as part of a libtissue server, and libtissue clients provide input data to the algorithm and response mechanisms which change the state of the monitored system. This client/server architecture separates data collection by the libtissue clients from data processing by the libtissue servers and allows for relatively easy extensibility and testing of algorithms on new data sources. The libtissue system is coded in C as a Linux shared library with client and server APIs, allowing new antigen and signal sources to be easily added to libtissue servers through the use of callbacks provided by these APIs. Because libtissue is implemented as a library, algorithms can be compiled and run on other researchers' machines with no modification. Clients and servers can potentially run on separate machines, for example a signal or antigen client may in fact be a remote network monitor.

AISs are implemented within a libtissue server as multi-agent populations of cells. Cells exist within an environment, called a tissue compartment, along with other cells, antigen and signals. The problem to which the algorithm is being applied is represented by libtissue as antigen and external signals. Clients in libtissue collect antigen and external signals and pass them to the libtissue server, which makes them available to the AISs. Cells express various repertoires of receptors and producers which allow them to interact with antigen and control other cells through signalling networks. Additionally, libtissue allows data on implemented algorithms to be collected and logged, allowing for experimental analysis of the system.

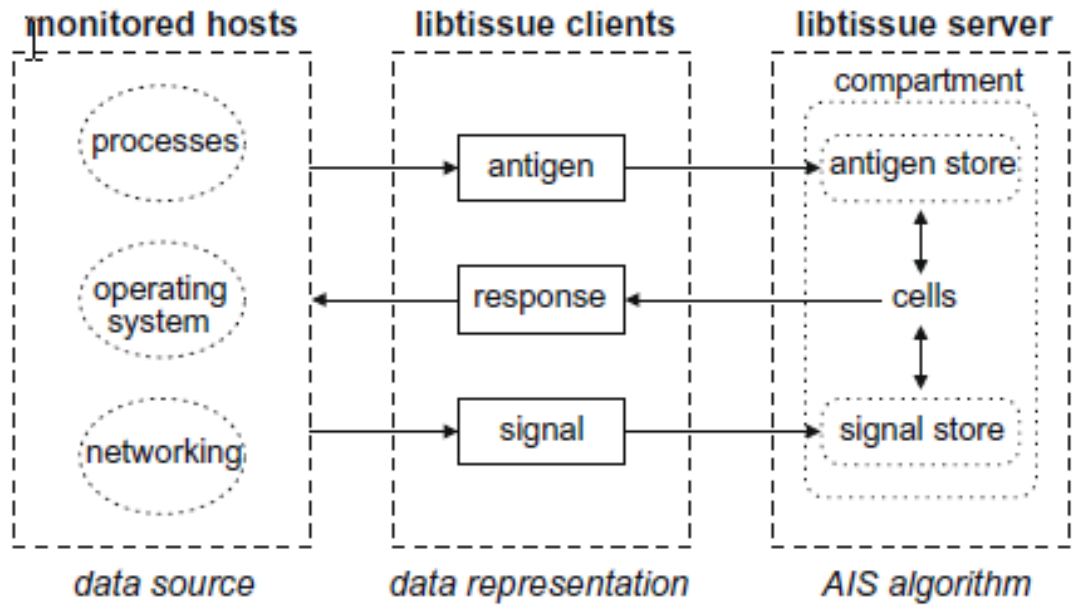

Fig. 1. The architecture of libtissue. Clients in libtissue monitor a host and provide input data to a libtissue server and AIS algorithm. Clients also allow algorithms to change the state of the monitored host.

Building an AIS using libtissue essentially involves creating a tissue compartment or compartments and populating these compartments with populations of different types of cell. Pseudocode for skeleton algorithm implemented in libtissue is given in Algorithm 1. First, a file containing all the parameters for the algorithm is read in. Next, the compartments and the maximum number of cells, antigen and signals the compartments can store are created (initialise tissue subroutine). Different types of cells are then initialised (create cells subroutine). They are initialised with different sets of producers and receptors, which determine their input and output capabilities, and the other cell types they can interact with. These cells are placed into tissue compartments as they are initialised. Usually, a function to log data periodically as the AIS is running, called a probe, is started by the user (initialise tissue probe subroutine). The probe is started after the compartments have been initialised and populated with cells. Finally, the libtissue scheduler (step tissue subroutine) is called periodically to update the tissue compartments and the cells they contain.

**Algorithm 1.** Pseudocode for a typical libtissue algorithm.

```
read parameter file
call initialise tissue
call create cells
call initialise tissue probe
for ever do
  call step tissue
  sleep for cell_update_rate
end for
subroutine initialise tissue do
  # max_cells, max_antigen, max_cytokines
  create tissue compartment to store cells, antigen and signals
  start tissue client thread
end subroutine
subroutine create cells do
  # create num_cells cells
  for each cell do
    create cell according to cell-specific parameters
    set cell cycle callback # cell_cycle_callback
    place cell at a random location in tissue compartment
  end for
end subroutine
subroutine initialise tissue probe do
  set tissue probe callback for tissue probe # probe_callback
  set time interval for tissue probe # probe_rate
  start tissue probe thread
end subroutine
subroutine cell cycle callback # cell_cycle_callback do
  # algorithm-specific cell controller
  process input from receptors and set producers accordingly
end subroutine
subroutine tissue probe callback # probe_callback do
  # algorithm-specific data logging routine
  write algorithm-specific data to log file
end subroutine
```

## 5. Process anomaly detection

We have used libtissue to implement several second generation AISs that are used for dynamic anomaly detection. One of these AISs will be described shortly, but first, in this section, we discuss the nature of the specific problem domain, intrusion detec- tion, on which we have tested our AISs.

Intrusion detection systems are designed to identify and pre- vent the misuse of individual computers and networks of comput- ers [21]. Such systems can be classified, based upon the analysis approach they employ, as either misuse detection or anomaly detection systems [9]. Misuse detection examines network and system activity for known misuses, usually through some form of pattern-matching algorithm. In contrast, anomaly detection sys- tems base their decisions on a profile of normal network or system behaviour, often constructed using statistical or machine learning techniques.

Each of these two approaches offers its own strengths and weaknesses. Misuse-based systems generally have quite low false positive rates but are unable to identify novel or obfuscated attacks, leading to high false negative rates. Anomaly-based systems, on the other hand, are able to detect novel attacks but currently produce a large number of false positives [3]. This stems from the inability of current anomaly-based techniques to cope adequately with the fact that in the real world normal, legiti- mate computer network and system usage changes over time, meaning that any profile of normal behaviour also needs to be dynamic.

Our work is aimed at developing an anomaly-based intrusion detection system which is able to cope with changing patterns of normal behaviour. An open problem with such sys- tems is the reduction of false positive rates while maintaining a high true positive rate [3]. Biological immune systems, which have to adapt to changing conditions over the lifetime of an organism, are an important source of inspiration when attempting to building artificial systems with the same proper- ties. Such systems are able to identify effectively anomalous events even though the normal state of the organism changes considerably as a result of environmental conditions and ageing.

A number of intrusion detection systems have been built around monitoring running processes to detect intrusions, termed process anomaly detection. In general, these collect information about a running process from a variety of sources, including from log files created by the process, or from other information gathered from the operating system. The idea is that by observing what the process is currently doing, for example by looking at its log files, we can tell whether the process is behaving normally or has been subverted by an attack.

While log files are an obvious starting point for such systems, and are still an important component in a holistic security approach, attacks may not cause any logging to take place, and so evade detection. Because of this, there has been a substantial amount of research into other data sources, usually collected by the operating system. Of these, system calls (syscalls) have been the most favoured approach. Syscalls are a low-level mechanism by which applications request system services such as peripheral I/O or memory allocation from an operating system. As a process runs it cannot usually directly access memory or hardware devices. Instead, the operating system manages these resources and pro- vides a set of functions, called syscalls, which processes can call to access these resources.

Furthermore, recent research [12,39] suggests that syscalls can be combined with other sources of information to increase the detection capabilities of syscall-based anomaly detection systems. In this respect, there is a convergence between intrusion detection and multi-sensor data fusion research. Our work is focussed on developing immune-inspired algorithms which use syscalls combined with resource usage statistics to decrease the false positive rate of anomaly detection systems. Resource usage statistics are indicators of process behaviour gathered at runtime, such as CPU, memory or file usage indicators. One of the important properties of second generation AISs is their use of multiple input data sources which reflect behaviour at a number of levels. Our idea is that resource usage statistics and other information provided by the operating system can be combined with syscall information to provide these multiple input data sources for second generation AISs. In such AISs, syscalls and re- source usage statistics form the antigen and external signals, respectively.

Table 2
Statistics for the six rpc.statd datasets gathered. For each monitored session, the table lists the total duration of the session (in seconds), the total number of antigen (i.e. syscalls) collected, the maximum number of antigen observed per second, the number of signals monitored, and the total number of signals collected.

| Session | Total time | Total antigen | Max rate | Num signals | Total signals |
|---|---|---|---|---|---|
| success1 | 55 | 1739 | 1102 | 2 | 474 |
| success2 | 36 | 1743 | 790 | 2 | 316 |
| failure1 | 54 | 518 | 405 | 2 | 461 |
| failure2 | 68 | 495 | 405 | 2 | 590 |
| normal1 | 38 | 434 | 405 | 2 | 334 |
| normal2 | 104 | 450 | 405 | 2 | 908 |

Data were collected on the behaviour of a number of servers under a range of normal and attack usage. In this paper we em- ploy a dataset generated by monitoring an RPC (Remote Proce- dure Call) statd server. Such a server is used by network file systems to determine when a computer has rebooted. The server was monitored under normal and attack conditions, and syscalls and two resource usage statistics (CPU and memory usage) were gathered to provide sources of antigen and external signals, respectively. Statistics for the six rpc.statd sessions are given in Table 2. The dataset is available online [23] and more informa- tion is given in [34].

## 6. The twocell algorithm

In this section a second generation AIS, twocell, that was implemented using libtissue is described. This algorithm utilis- es several important properties of second generation AISs, such as multiple cell types, multi-level input signals and internal signals, and shows how these properties can be implemented in libtissue.

The twocell algorithm is a second generation AIS, implementing aspects of biological innate and adaptive immunity. In particular, twocell models innate immune system dendritic cells (DCs) and adaptive immune system T cells. The cells in twocell, shown schematically in Fig. 2, are of two types, labelled Type 1 and Type 2, and each type has different receptor and producer repertories. Pseudocode for twocell is given in Algorithm 2. Receptors allow cells to sense information from different sources. Antigen receptors allow structural information about the problem to be sensed and cytokine receptors allow behavioural information to be sensed. Each cell type has a different cell cycle callback which determines how the information received through its receptors is integrated. This information in turn determines the function of the cell and the signals it produces.

Algorithm 2. Pseudocode for the twocell algorithm.

```
subroutine dc cell cycle callback # type 1 cell
  if signal level in tissue compartment has increased then
    for all antigen producers # num_antigen_producers_1 do
      set action time of antigen producer to
   antigen_producer_action_time
    end for
  end if
  ifsignal level in tissue compartment has decreased bf then
    for all antigen producers bf do
      set action time of antigen producer to 50% of current action
   time
    end for
  end if
end subroutine
subroutine tc cell cycle callback # type 2 cell bf do
  if cell iterations >¼ cell_lifespan_2 bf then
    replace cell with a new tc
    return
  end if
  for all vr receptors # num_vr_receptors_2 bf do
    if vr receptor activated then
      write matched antigen to log file
    end if
  end for
end subroutine
```

Type 1 cells are designed to emulate two key functions of biological DCs: antigen and signal processing. For antigen processing, each Type 1 cell is equipped with a number of antigen receptors and producers. Antigen is collected by Type 1 cells using antigen receptors and presented to Type 2 cells using antigen producers. This allows Type 1 cells to aggregate antigen into temporally-re- lated groups. A cytokine receptor allows Type 1 cells to respond to the value of a signal in the tissue compartment.

Type 2 cells emulate three of the functions of biological T cells: cellular binding, antigen matching and antigen response. Each Type 2 cell has a number of cell receptors specific for Type 1 cells, VR receptors to match antigen, and a response producer which is triggered when antigen is matched. Type 2 cells also maintain one internal cytokine, an integer which is incremented every time a match between an antigen producer and VR receptor occurs. If the value of this cytokine is still zero, that is no match has oc- curred, after a certain number of cycles then the values of all of the VR receptor locks on the cell are randomised.

A tissue compartment is created and populated with a number of Type 1 and 2 cells. Antigen and signals in the compartment are set by libtissue clients based on the syscalls a process is making and its CPU usage. Type 1 cells ingest antigen through their antigen receptors and present it on their antigen producers. The period for which the antigen is presented is determined by a signal read by a cytokine receptor on these cells, and so can be made dependant upon CPU usage.

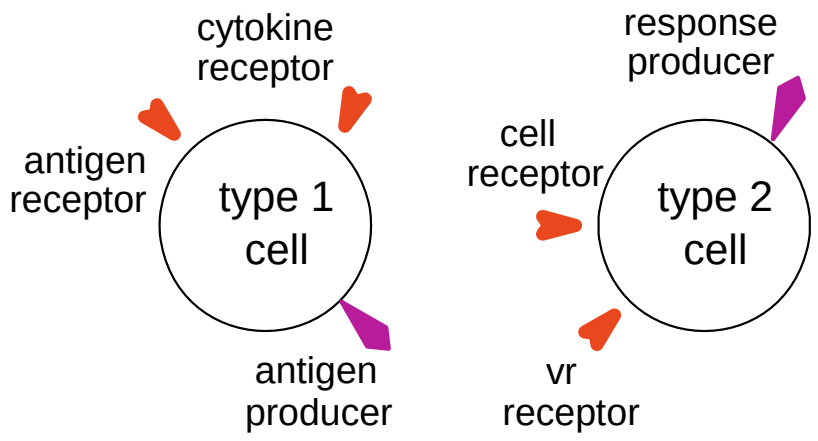

Fig. 2. The two different cell types implemented in twocell. Type 1 cells ingest antigen through antigen receptors and display antigen on their surface via antigen producers. They also respond to an external signal through a cytokine receptor, which determines the amount of time antigen is presented for. Type 2 cells bind with Type 1 cells via cell receptors and then match antigen presented on Type 1 cells with VR receptors. If a match occurs Type 2 cells produce an alert through their response producers.

Table 3
The libtissue parameter settings used for twocell.

| | |
|---|---|
| max_antigen | 1000 |
| max_cytokines | 0 |
| max_cells | 100 |
| cell_update_rate (1s) | 100,000 |
| antigen_multiplier | 10 |
| num_cells 1 | 50 |
| num_antigen 1 | 100 |
| num_antigen_receptors 1 | 10 |
| num_antigen_producers 1 | 10 |
| antigen_producer_action_time | 10 |
| num_cells 2 | 50 |
| cell_lifespan 2 | 100 |
| num_cell_receptors 2 | 2 |
| num_vr_receptors 2 | 20 |
| num_response_producers 2 | 1 |
| probe_rate (1s) | 1,000,000 |

Table 4
The naive syscall policy and the average twocell policy generated from the normal1 and normal2 datasets. The first column lists the names and numbers (in brackets) of syscalls that are permitted in the naive policy. The second column gives the number of times each syscall appears in both datasets. The third column gives the mean number of times a syscall appears in a twocell policy over 20 runs. The fourth column gives the standard deviations of these means, and the fifth column gives the coefficient of variation.

| Syscall | Frequency | Mean | sd | cv |
|---|---|---|---|---|
| chdir(12) | 2 | 0.07 | 0.26 | 371 |
| execve(11) | 2 | 0.07 | 0.26 | 371 |
| personality(136) | 2 | 0.07 | 0.34 | 485 |
| setsid(66) | 2 | 0.07 | 0.34 | 485 |
| fork(2) | 2 | 0.10 | 0.37 | 370 |
| write(4) | 2 | 0.10 | 0.37 | 370 |
| send(309) | 2 | 0.15 | 0.56 | 373 |
| time(13) | 2 | 0.15 | 0.40 | 266 |
| fstat64(197) | 2 | 0.17 | 0.52 | 305 |
| lseek(19) | 2 | 0.17 | 0.42 | 247 |
| fsync(118) | 2 | 0.25 | 0.80 | 365 |
| getrlimit(191) | 2 | 0.28 | 0.67 | 320 |
| listen(304) | 2 | 0.28 | 0.63 | 239 |
| select(142) | 3 | 0.57 | 1.48 | 225 |
| gettimeofday(78) | 4 | 0.50 | 0.85 | 276 |
| getsockname(306) | 4 | 0.53 | 1.47 | 170 |
| _exit(1) | 4 | 0.55 | 1.38 | 277 |
| uname(122) | 4 | 0.75 | 1.91 | 250 |
| stat(106) | 4 | 0.80 | 2.58 | 259 |
| connect(303) | 5 | 1.60 | 2.48 | 254 |
| getdents(141) | 8 | 0.20 | 0.73 | 322 |
| mprotect(125) | 8 | 0.47 | 1.30 | 185 |
| poll(168) | 8 | 0.90 | 1.67 | 224 |
| sendto(311) | 9 | 0.95 | 2.13 | 225 |
| recvfrom(312) | 9 | 2.45 | 3.68 | 233 |
| rt_sigaction(174) | 10 | 0.97 | 2.19 | 155 |
| getpid(20) | 10 | 1.60 | 2.28 | 142 |
| fcntl(55) | 12 | 1.18 | 2.76 | 268 |
| bind(302) | 12 | 1.68 | 4.51 | 200 |
| munmap(91) | 15 | 1.88 | 3.77 | 225 |
| brk(45) | 16 | 2.25 | 3.78 | 168 |
| fstat(108) | 23 | 2.33 | 4.45 | 229 |
| ioctl(54) | 24 | 2.73 | 4.67 | 190 |
| socket(301) | 25 | 3.10 | 4.97 | 150 |
| old_mmap(90) | 27 | 1.90 | 4.29 | 171 |
| read(3) | 27 | 2.25 | 5.17 | 160 |
| open(5) | 30 | 5.95 | 7.75 | 130 |
| close(6) | 557 | 19.43 | 27.03 | 139 |

Type 2 cells attempt to bind with Type 1 cells via their cell receptors. If bound, VR receptors on these cells interact with anti- gen producers on the bound Type 1 cell. If an exact match between a VR receptor lock and antigen producer key occurs, the response producer on Type 2 cells produces a response, in this case a log en- try containing the value of the matched receptor and indicating that the syscall is permitted.

## 7. twocell experiments

The behaviour of twocell was examined in an number of experiments using the rpc.statd dataset, described in Section 5 above. The first experiment looks at a number of twocell runs, while the second takes one run and examines it more closely. The third evaluates the performance of a syscall policy generated by twocell. During these experiments, to more clearly understand the dynamics of twocell, the cytokine receptor on Type 1 cells is disabled, thus making twocell unresponsive to the CPU usage external signal. The final experiment returns to the question of signals and compares the effect the addition of the signal, i.e. CPU usage, has on the dynamics of twocell. The parameters given in Table 3 were used for all experiments, which were carried out on a 2 GHz AMD64 Turion laptop running Debian Linux. Runs used on average around 1%, and never more than 3%, of the available CPU resources.

In experiments it is important to have a baseline with which to compare algorithmic performance. In terms of syscall policies such a baseline can be generated, and is here termed a naive policy. A na-ive syscall policy is generated for a process, such as rpc.statd, by recording the syscalls it makes under normal usage, as in the normal1 and normal2 datasets. A permit policy statement is then created for all syscalls seen. This baseline whitelist approach is not too unrealistic when compared to how current systems such as systrace automatically generate a policy. The first column of Table 4 shows the permitted syscalls (syscall number given in brackets) in such a naive policy generated from the normal1 and normal2 datasets. The frequency with which each syscall was observed, combined over the two datasets, is given in the second column, as this will be useful for further analysis. Fig. 3a shows the syscall number plotted against its frequency.

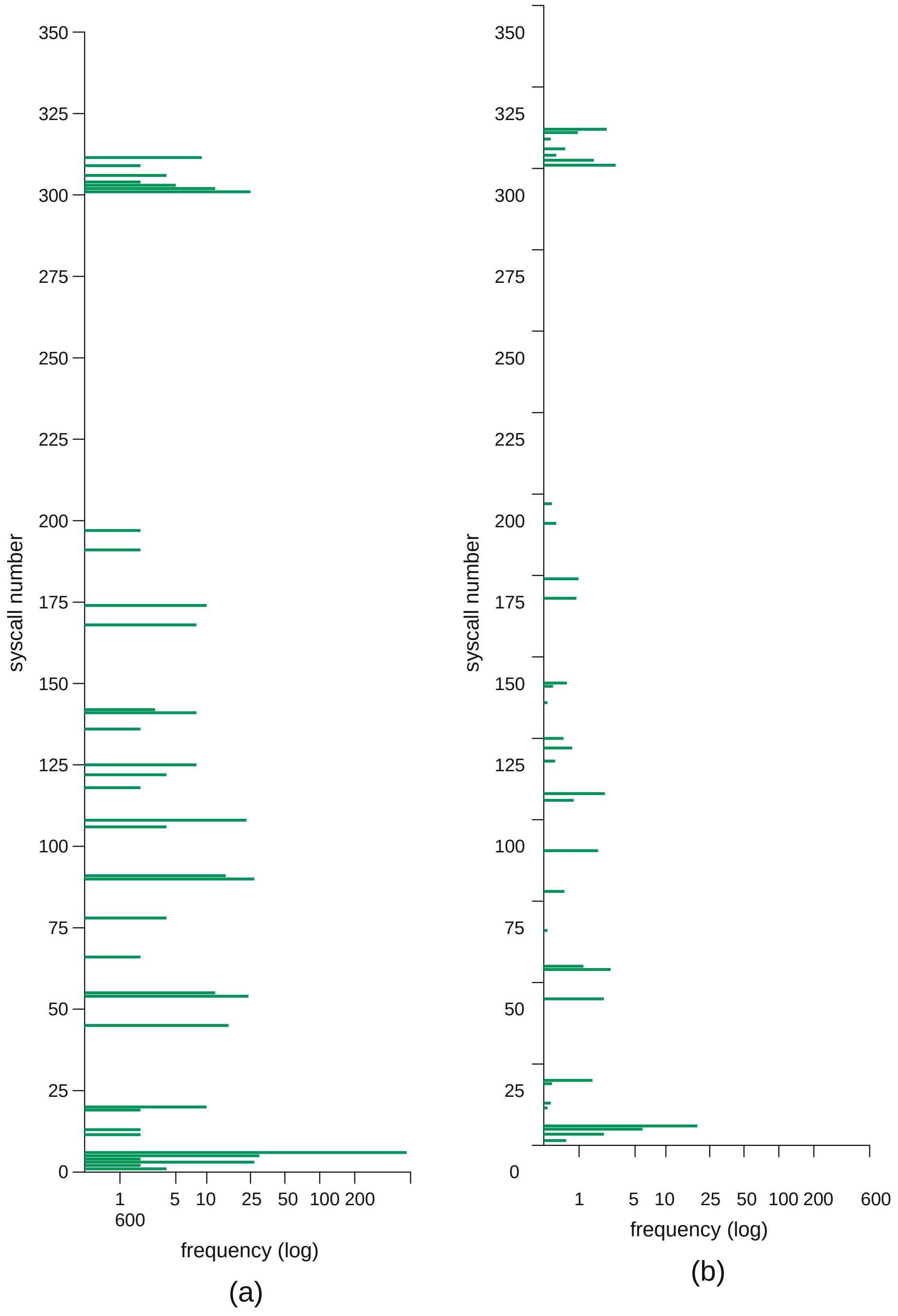

Fig. 3. The frequencies of the syscalls seen in the normal1 and normal2 datasets (a), and the frequencies of the syscalls produced over the 40 runs of twocell (b) on the same datasets.

Similarly to a naive policy, one way in which twocell can be used to generate a syscall policy is by running it with normal usage data during a training phase. During this phase, responses made by Type 2 cells are recorded. At the end of the training phase, a syscall policy is created by allowing only those syscalls responded to, and denying all others. Since interactions in libtissue are stochastic, looking at the average results over a number of runs is necessary to understand the behaviour of implemented algorithms. A script starts the twocell server and then after 10 s starts the tcreplay client and replays a dataset in realtime. The twocell server continues running for a further minute after replay had finished. This process is repeated 20 times for both the normal1 and normal2 datasets, yielding 40 individual syscall policies. A single twocell policy is then generated by allowing all syscalls which are permitted in any of the 40 individual policies.

The null hypothesis ð$H_0$Þ for this experiment is that there is no difference in the response of twocell for syscalls with different frequencies. The alternative hypothesis ð$H_1$Þ is that twocell responds differently depending on the frequency of the syscall. The second column of Table 4 and Fig. 3b show the frequency of each syscall. The third column of Table 4 and Fig. 3b show the mean frequency with which each syscall appears in a twocell policy. We found that all of the 38 syscalls that occur are also permitted in the twocell policy. The Spearman rank correlation coefficient was calculated in the standard way for the distributions in these two columns, and was found to be q ¼ 0:9285, which is larger than the critical value for q at p < 0:001. Therefore, the null hypothesis ð$H_0$Þ is false, and twocell responses are correlated to the frequency of the syscalls. Standard deviations, given in the fourth column of Table 4, appear at first to show an increasing amount of noise for high-frequency syscalls. However, examination of the coefficient of variation for each syscall, given in the last column of Table 4, shows that there is in fact more variation in the frequencies of response to the lower frequency syscalls.

Table 5
The syscall policy generated by twocell and the frequency of response for each syscall for the normal2 dataset.

| Syscall | Frequency |
|---|---|
| gettimeofday(78) | 1 |
| listen(304) | 1 |
| send(309) | 1 |
| select(142) | 2 |
| poll(168) | 3 |
| recvfrom(312) | 8 |
| fcntl(55) | 9 |
| fstat(108) | 9 |
| open(5) | 22 |
| close(6) | 34 |

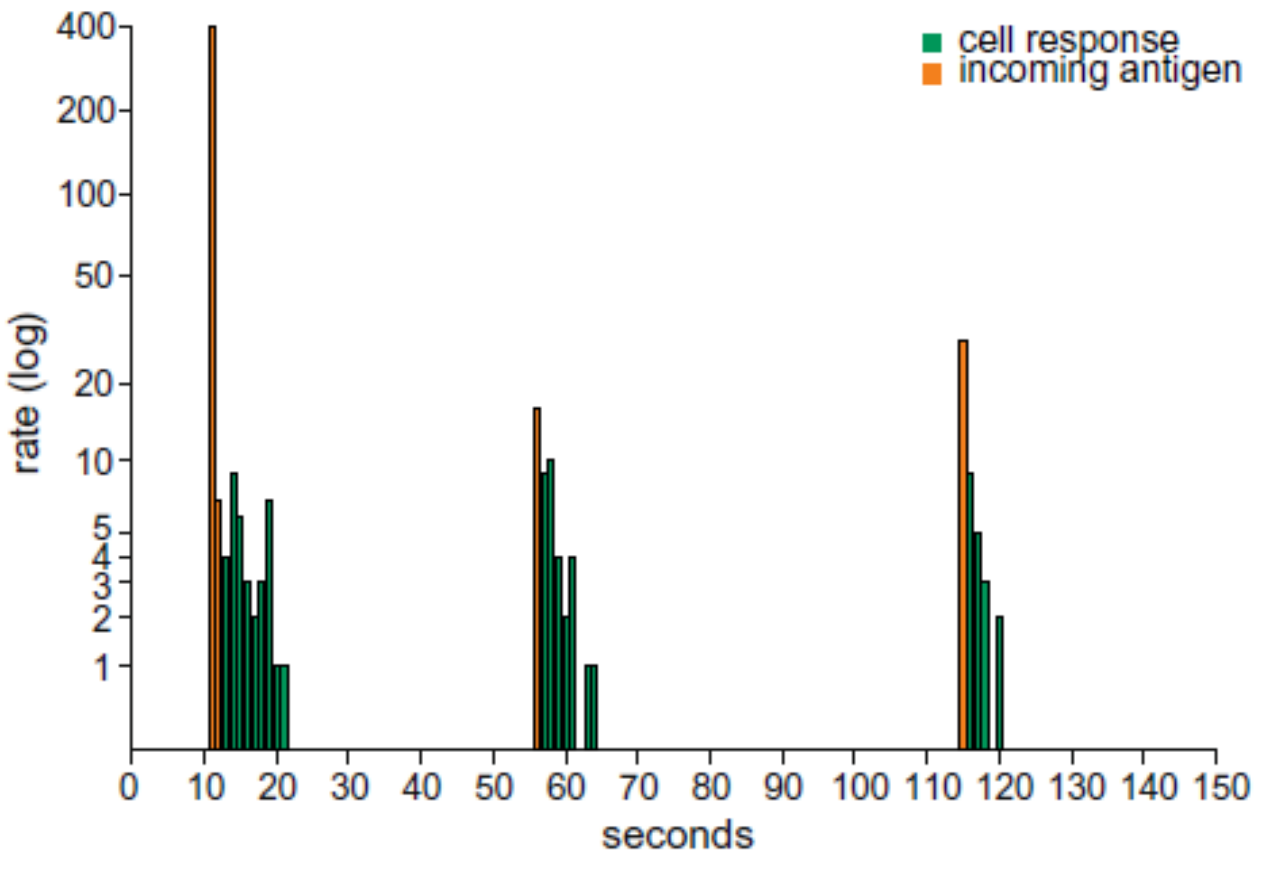

Fig. 4. The rate of incoming antigen and corresponding cell response rates produced by twocell for the normal2 dataset. The amount of incoming antigen is shown in orange (light grey), and the number of responses generated by Type 2 cells in green (dark grey).

The previous experiment shows that the twocell algorithm has the property of responding in a selective way to input data based on the frequency at which an input data item occurs. In order to examine more closely how twocell responds, a single run of the twocell algorithm is now analysed. Following the same general procedure as the previous experiment, twocell is run once with the normal2 dataset. The resulting policy is shown in Table 5, along with the frequencies with which the permitted syscalls are responded to. During the run, the time at which a Type 2 cell produces a response to a particular syscall is also recorded, and the rate at which these responses occur is plotted in Fig. 4. The rate of incoming syscalls is also plotted for comparison. This figure clearly shows a correlation between the rate of incoming syscalls and the rate of responses produced by Type 2 cells. Cells initially do not produce any response until syscalls occur, and then produce a burst of responses for a relatively short period before settling down to an unresponsive state once again. This is to be expected, as antigens enter and are passed through twocell until their eventual destruction after being presented on Type 1 cell antigen producers.

For the same run, the individual receptors expressed by Type 2 cells can also be examined. Fig. 5 shows the repertoire of VR receptors expressed by all 50 Type 2 cells during the run. A libtissue probe periodically recorded the syscall values expressed by the VR receptors on all of the Type 2 cells. A point is plotted in Fig. 5 if the syscall was being expressed during that period. Points for the ten syscalls that twocell responded to (see Table 5) are highlighted. As expected, due to the limited lifespan of unmatched Type 2 cells, set by the cell_lifespan parameter, and after which the cell's VR receptor is randomised, many bursts of around 10 s of expression of VR receptors specific for a given syscall are seen. Once a VR receptor matches, and a response and permit policy is therefore produced for that syscall, the cell stops randomising its receptors. This can be observed from the continuous horizontal lines in Fig. 5 for the 10 highlighted syscalls.

An example is now given of how the classification accuracy and error of a libtissue algorithm can be evaluated. In terms of syscall policies, a particular policy can be considered successful in relation to the number of normal syscalls it permits versus the number of attack syscalls it denies. The naive policy and average twocell policy generated from datasets normal1 and normal2 in the first experiment above are evaluated in such a way.

The number of syscalls both policies permit and deny when applied to the four datasets in the attack and failed groups is recorded. For each dataset, Table 6 shows the percentages of syscalls permitted by the naive and twocell policies. From the results, the tendency of the naive policy is to permit the vast majority of syscalls, whether attack related or not. The twocell generated policy behaves much more selectively, denying a slightly larger proportion of syscalls in the success1 and success2 datasets than it permits. For the failure1 and failure2 datasets the converse is true.

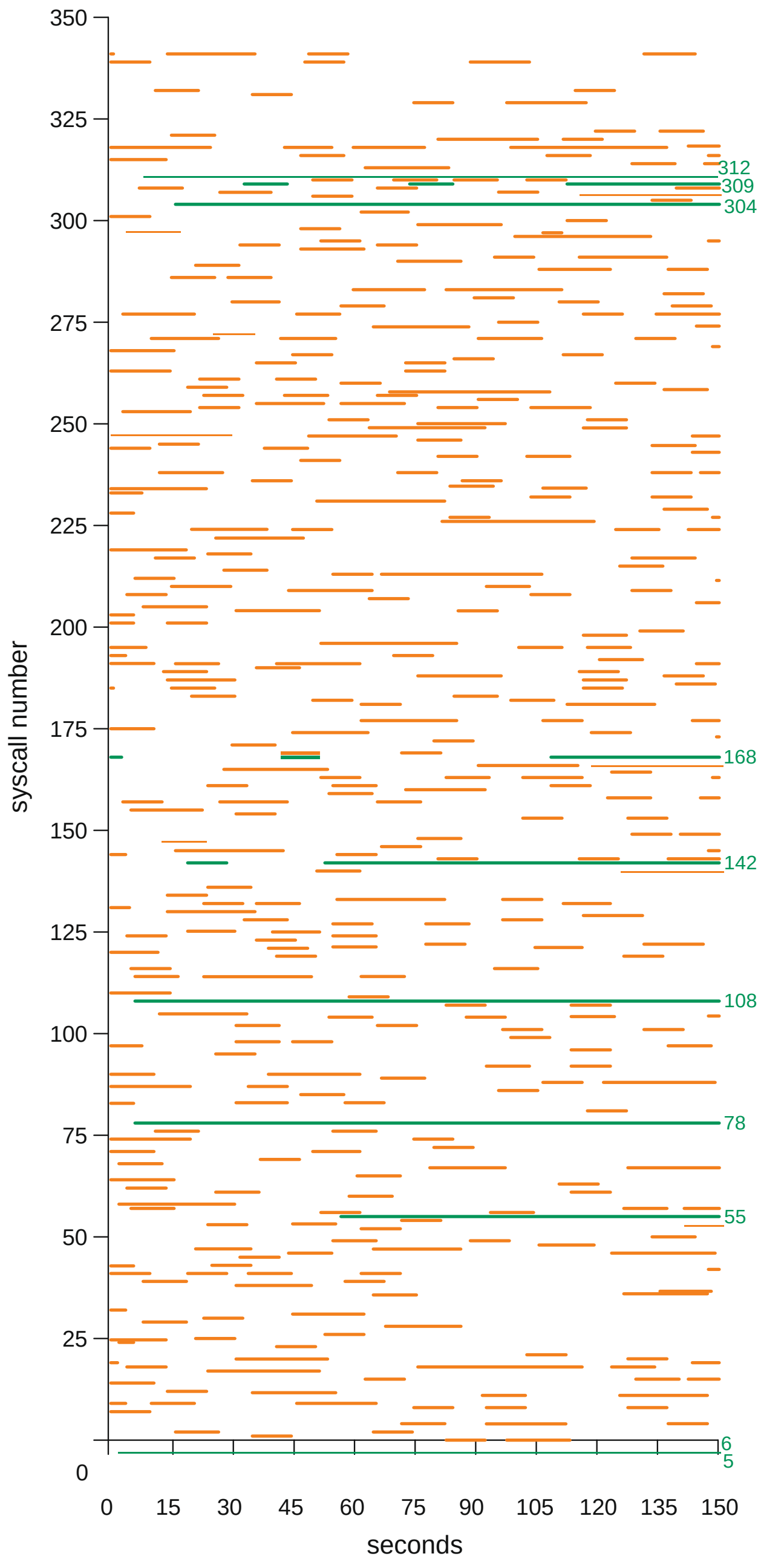

Fig. 5. The VR receptor repertoire expressed by Type2 cells generated by twocell for the normal2 dataset. An orange (light grey) point is plotted for the corresponding syscall whenever a Type 2 cell with a VR receptor specific for the syscall is present. Green (dark grey) points indicate that the Type 2 cell also produced a response to the syscall.

Table 6
Performance of a naive policy and a twocell policy generated from the normal2 dataset. The naive policy permits the majority of syscalls in all four datasets, and denies only a small number of syscalls. For both of the success datasets, twocell twocell permits around two thirds of syscalls, and denies one third.

| Dataset | naive | | twocell | |
|---|---|---|---|---|
| | Permitted (%) | Denied (%) | Permitted (%) | Denied (%) |
| success1 | 91 | 9 | 48 | 52 |
| success2 | 91 | 9 | 48 | 52 |
| failure1 | 100 | 0 | 70 | 30 |
| failure2 | 100 | 0 | 69 | 31 |

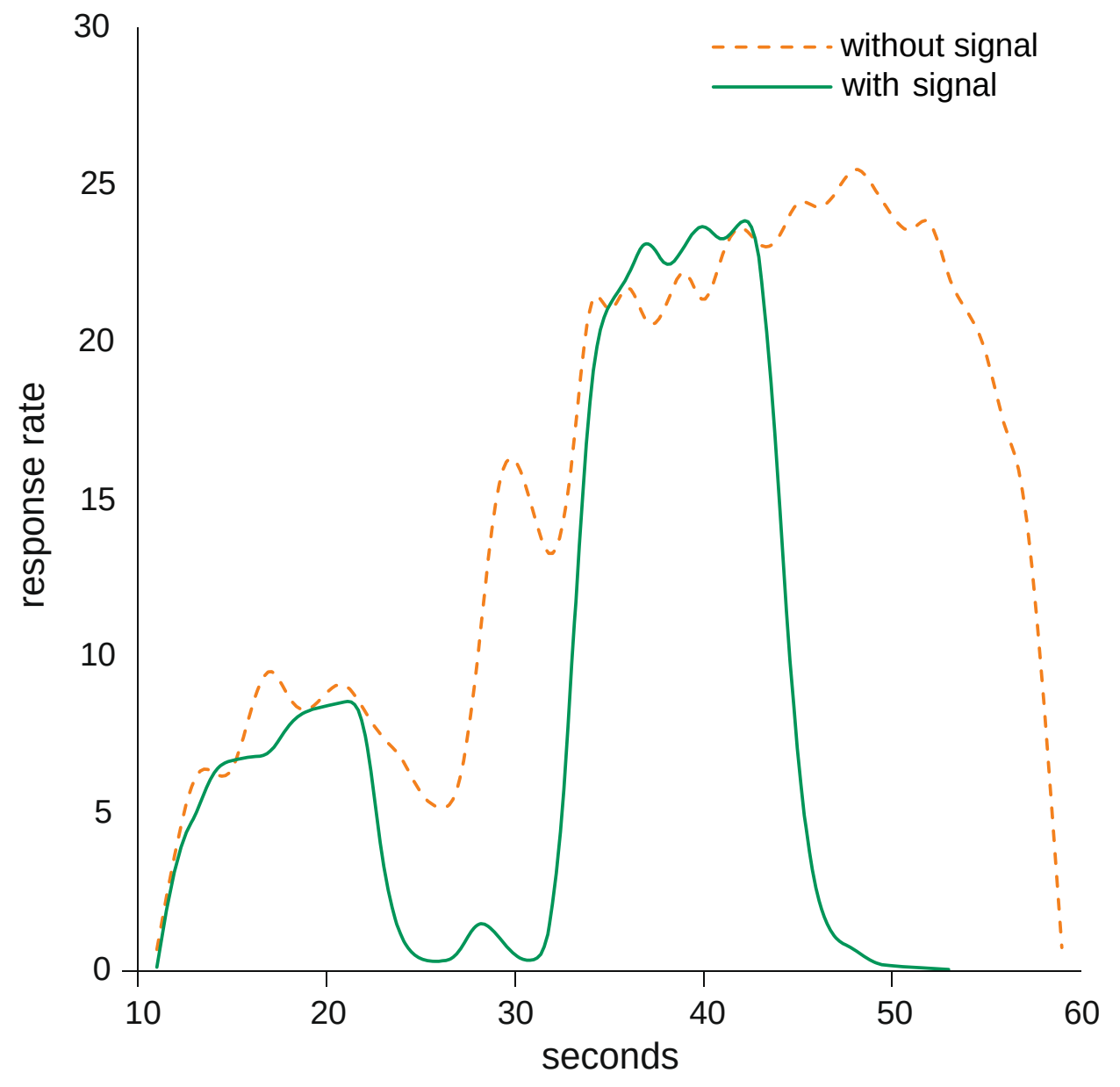

Fig. 6. The mean response rates produced by Type 2 cells of the twocell algorithm with and without a signal for 20 runs on the success2 dataset.

The previous experiments have all used the twocell algorithm with the cytokine receptors of Type 1 cells disabled. This was necessary to gain an initial understanding of the dynamics of twocell. This final experiment now examines how the addition of a context signal changes the dynamics of the algorithm. The hypothesis for this experiment is that the addition of a context signal to twocell does not change the response in terms of Type 2 cells (the null hypothesis $H_0$). The alternative hypothesis ð$H_1$Þ is that the addition of a context signals changes the response of twocell in terms of Type 2 cells.

When enabled, the cytokine receptor on a Type 1 cell controls the action_time parameter of antigen producers on these cells as follows. The action_time parameter is initialised to a value of 100. If there is no change in the signal, CPU usage in this case, then the action time stays the same. If CPU usage has decreased, the action time is reduced by 50%, and if it has increased, the action time is reset to 100. The twocell algorithm with its cytokine receptor en- abled is run 20 times on the success2 dataset and the responses it produces are recorded. For a fair comparison, the mean action time observed on antigen producers over all of the runs, 28.57 in this case, is calculated and the twocell algorithm without signals is run 20 times on the same dataset with the action time of its anti- gen producers set to 29.

Fig. 6 shows bspline curves fitted to the mean response rates of twocell with and without a signal over the 20 runs. The results show that the response time of twocell with a signal is much more tightly controlled, with responses starting and dropping off more rapidly and lasting for a shorter duration in total. The Spearman rank correlation coefficient was calculated in the standard way for the distributions of the response rates with and without a signal. A value of q ¼ 0:9076 was obtained, which is larger than therefore false and there is a significant change in Type 2 cell responses when a context signal is added to twocell. This can be explained in light of the incoming data, and from the action of the cytokine receptor, which causes a sudden rise and quick decreases in the action time of the antigen producers on Type 1 cells based on

## 8. Conclusions

This paper has shown how multi-level data fusion mechanisms seen in the biological immune system can be used to build AISs. These AISs are represented as populations of autonomous agents. This representation works well in that it is fairly straightforward to transfer biological experimental models of the immune system into AISs. The use of multi-level information sources in these AISs shows how a real-world problem can be framed to reflect the envi- ronment of the biological immune system seen as a combination of both the innate and the adaptive immune systems. This problem representation proved to be a good source of multi-level input data for second generation AISs. The provision of signal and antigen receptors by libtissue was shown to be useful in providing agents with access to the sources of multi-level input data avail- able from this problem representation.

The immune system provides a good example of a parallel and distributed biological information fusion processes. Second generation AISs modelling some of these processes have been used to build anomaly detectors with low false positive rates [34]. Central to second generation AISs is the idea of a multi-level representation of the problem as the environment of the AIS. Coupled with this multi-level problem representation is the representation of the AIS as composed of populations of agents of multiple types. These agents interact with each other and the environment to establish a homeostatic balance of cell populations. The mecha- nisms of biological information fusion modelled by second gener- ation AIS might also prove useful in multi-sensor information fusion.